\documentclass[10pt,twocolumn,letterpaper]{article}

\usepackage[pagenumbers]{cvpr} % To force page numbers, e.g. for an arXiv version
\usepackage{multirow}
\usepackage{pifont}

\usepackage[dvipsnames]{xcolor}

\usepackage{graphicx}
\usepackage{amsmath}
\usepackage{amssymb}
\usepackage{booktabs}
\usepackage{multirow}
\usepackage{bbding}
\usepackage{bm}
\usepackage{tabularx}
\usepackage[ruled,vlined]{algorithm2e}

\definecolor{cvprblue}{rgb}{0.21,0.49,0.74}
\usepackage[pagebackref,breaklinks,colorlinks,citecolor=cvprblue]{hyperref}

\usepackage[capitalize]{cleveref}

\def\paperID{5} % *** Enter the Paper ID here
\def\confName{CVPR}
\def\confYear{2024}

\definecolor{commentcolor}{RGB}{110,154,155}   % define comment color
\newcommand{\PyComment}[1]{\ttfamily\footnotesize\textcolor{commentcolor}{\# #1}}  % add a "#" before the input text "#1"
\newcommand{\PyCode}[1]{\ttfamily\footnotesize\textcolor{black}{#1}} % \ttfamily is the code font

\title{{\em i-MAE}: Are Latent Representations in Masked Autoencoders Linearly Separable?}

\author{Kevin Zhang$^{\dagger,*}$, Zhiqiang Shen$^{\S,}$\thanks{The two authors have equal contribution to this work. Zhiqiang Shen is the corresponding author. Code is available on \href{https://github.com/VILA-Lab/i-mae}{GitHub}.}\\
$^\dagger$Peking University  \ $^\S$Mohamed bin Zayed University of AI  
}

\begin{document}

\maketitle
\begin{abstract}
\vspace{-0.05in}
Masked image modeling (MIM) has been recognized as a strong self-supervised pre-training approach in the vision domain. However, the mechanism and properties of the learned representations by such a scheme, as well as how to further enhance the representations are so far not well-explored.  
In this paper, we aim to explore an interactive Masked Autoencoders (i-MAE) framework to enhance the representation capability from two aspects: {\bf (1)} employing a two-way image reconstruction and a latent feature reconstruction with distillation loss to learn better features; {\bf (2)} proposing a semantics-enhanced sampling strategy to boost the learned semantics in MAE. Upon the proposed i-MAE architecture, we can address two critical questions to explore the behaviors of the learned representations in MAE: {\bf (1)} Whether the separability of latent representations in Masked Autoencoders is helpful for model performance? We study it by forcing the input as a mixture of two images instead of one. {\bf (2)} Whether we can enhance the representations in the latent feature space by controlling the  degree of semantics during sampling on Masked Autoencoders? To this end, we propose a sampling strategy within a mini-batch based on the semantics of training samples to examine this aspect. 
Extensive experiments are conducted on CIFAR-10/100, Tiny-ImageNet and ImageNet-1K datasets to verify the observations we discovered. Furthermore, in addition to qualitatively analyzing the characteristics of the latent representations, we examine the existence of linear separability and the degree of semantics in the latent space by proposing two evaluation schemes. The surprising and consistent results across the qualitative and quantitative experiments demonstrate that i-MAE is a superior framework design for understanding MAE frameworks, as well as achieving better representational ability.
\end{abstract}    
\vspace{-0.1in}
\section{Introduction}
\label{sec:intro}
\vspace{-0.05in}

 \begin{figure}
    \centering
    \includegraphics[width=0.98\linewidth]{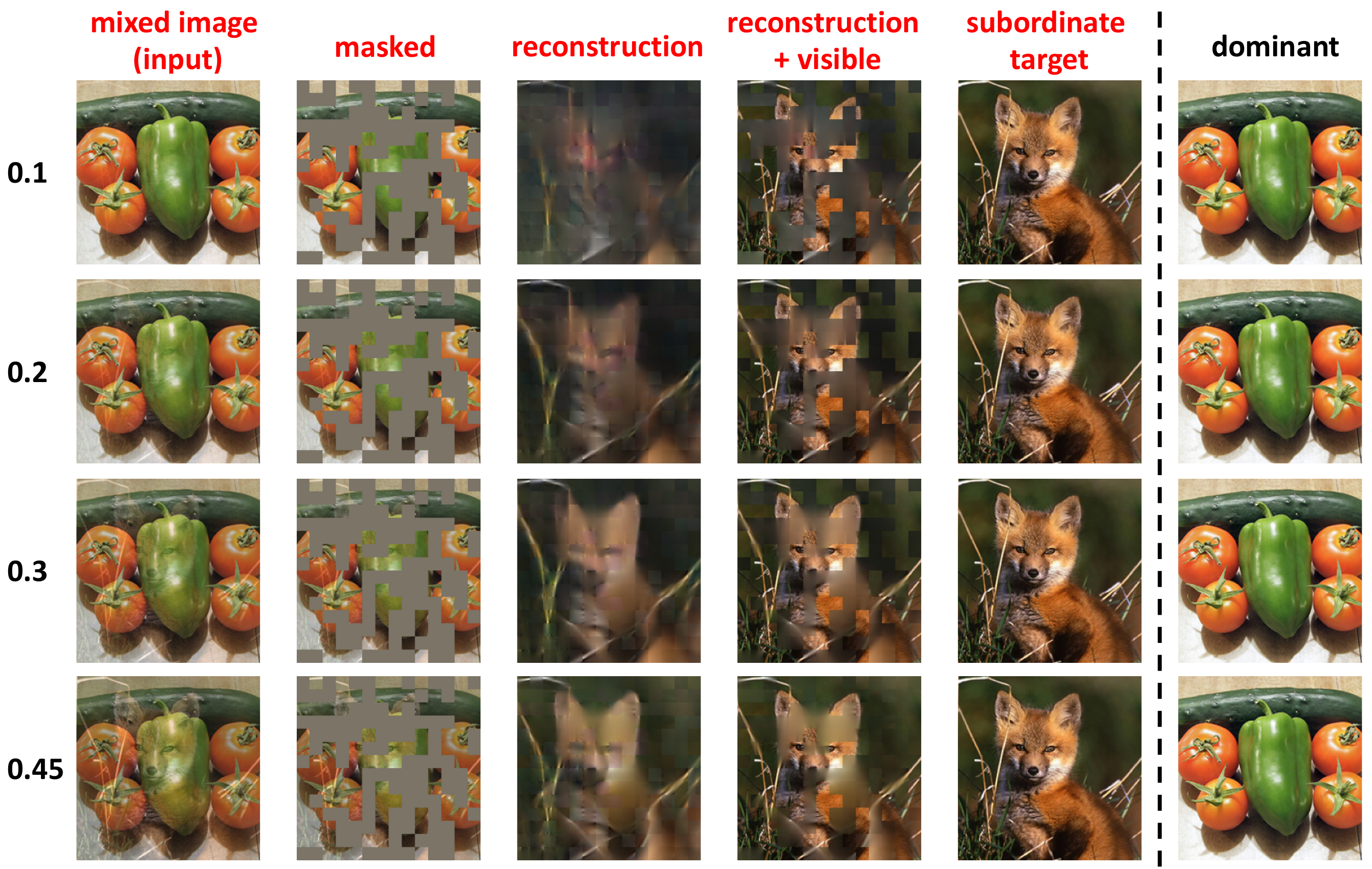}
    \vspace{-0.12in}
    \caption{Reconstruction results from the subordinate branch of {\em i-MAE} on ImageNet-1K validation images with different mixing coefficients ${\bm \alpha}$ (listed on the left). i-MAE is pre-trained with linearly mixed input reconstruction loss on both inputs. Visually, i-MAE predictions reflect features of ${\textbf I_s}$ even at low mixture coefficients and still reconstruct the subordinate image well, whereas at 0.45 which is more challenging, the reconstructions show elements like colors from the dominant image but the content still matches the target. More visualizations are provided in Appendix.}
    \label{fig:ImageNet}
   \vspace{-0.15in}
\end{figure}

Self-supervised learning aims to learn representations from abundant unlabeled data for benefiting various downstream tasks. Recently, many self-supervised approaches have been proposed in the vision domain, such as pre-text based methods~\cite{doersch2015unsupervised,zhang2016colorful,gidaris2018unsupervised}, contrastive learning with siamese networks~\cite{oord2018representation,he2020momentum,chen2020simple,henaff2020data}, masked image modeling (MIM)~\cite{he2022masked,bao2021beit,Xie2021SimMIMAS, wei2022masked}, etc. Among them, MIM has shown a preponderant advantage in finetuning performance, and the representative method Masked Autoencoders (MAE)~\cite{he2022masked} has attracted much attention in the field. A natural question is then raised: {\em Where are the benefits of the finetuning transferability to downstream tasks from in MAE-based training?} This motivates us to develop a framework to shed light on the reasons of MAE's superior latent representations. Further, we can explore the feasibility of leveraging the discoveries to continue enhancing the representation ability of MAE. As the understanding of MAE framework remains under-studied, it is crucial to explore it more in a specific and exhaustive way for better performance.

Intuitively, a good representation should be separable and contain enough semantics from its input, so that it can have a qualified ability to distinguish different classes with better performance on downstream tasks. This inspires us to question ``can we utilize the idea of separability and more semantics to enhance the representation capability for Masked Autoencoders?'' This is in general difficult as how to evaluate the separability and the degree of semantics on the latent features is not clear thus far. Moreover, the success of an Autoencoder {\em compressing} the information from input by reconstructing itself has been well-recognized only in practice; the explanation and interpretability of the features learned from such approaches is still under-explored.

To address the difficulties of identifying separability and semantics in latent features, we first propose a novel framework, i-MAE, upon vanilla MAE. It consists of a mixture-based masked autoencoder branch for disentangling the mixed representations by linearly separating two different instances, and a pre-trained vanilla MAE as guidance for distilling the disentangled representations. An illustration of the overview framework architecture is shown in Fig.~\ref{fig:framework}. This framework is designed to answer two interesting questions: {\bf (1)} Are the latent representations in Masked Autoencoders {\em linearly separable}? More importantly, how can we utilize this property to learn stronger and better features? {\bf (2)} Whether we can enhance the representations for Masked Autoencoders by leveraging {\em more semantics} through a sampling strategy within a mini-batch? 
We attribute the superior representation capability of MAE to its learned separable features for downstream tasks with enough semantics.

In addition to qualitative studies, we develop two quantitative evaluation schemes to address the two questions quantitatively. In the first evaluation, we employ several distance measurements of root $\ell_2$, {\em R-Squared} value after regression and {\em {cosine}}-similarity from high-dimensional Euclidean spaces to measure the similarity between i-MAE's disentangled feature and the ``ground-truth'' feature from pre-trained MAE on the same image. In the second evaluation, we control different ratios of semantic classes as a mixture within a mini-batch and evaluate the finetuning and linear probing results of the model to reflect the learned semantic information. More details are in Sec.~\ref{linear_method} and \ref{sema_method}.

We conduct extensive experiments on different scales of datasets: CIFAR-10/100, Tiny-ImageNet and ImageNet-1K to demonstrate the effectiveness of i-MAE with better capability, studying the linear separability and the degree of semantics in the latent representations. We further provide both qualitative and quantitative results to explain our observations and discoveries. 
The characteristics we observed in latent representations according to our proposed i-MAE framework are: {\bf(1)} i-MAE learned feature representation has great linear separability for its input data, which can be beneficial for linear probing and finetuning tasks. {\bf(2)} Though the training scheme of MAE is different from instance classification pre-text in contrastive learning, its representation still encodes sufficient semantic information from input data. Moreover, {\em mixing the same-class images as the input training samples substantially improves the quality of learned features}. {\bf(3)} We can reconstruct the individual images from a mixture with i-MAE effortlessly, even if it is the subordinate part. To the best of our knowledge, this is the pioneering study to explicitly explore the separability and semantics of a mixed MAE scheme with extensive well-designed qualitative and quantitative experiments.

Our contributions in this work are three-fold:
\begin{itemize}[leftmargin=0.15in]
\addtolength{\itemsep}{-0.0in}
\item We propose an {\em i-MAE} framework with two-way image reconstruction and latent feature reconstruction with a distillation loss to boost the representation capability inside the MAE framework and explore the understanding of mechanisms and properties of learned representations.

\item We evaluate our dual-reconstruction framework and find better linear separability of features which are more interpretable. We further introduce a semantics-enhancement sampling strategy, a straightforward yet effective scheme to increase the quality of learned features. 

\item We conduct extensive experiments on various scales of datasets: CIFAR-10/100, Tiny-ImageNet and ImageNet-1K, and we provide sufficient qualitative and quantitative results to verify the effectiveness of proposed framework. 
\end{itemize}
 
\section{Related Work}

\noindent{\textbf{Masked image modeling.}} Motivated by masked language modeling's success in language tasks~\cite{bert, gpt1}, Masked Image Modeling (MIM) in vision learns representations from images corrupted by masking. State-of-the-art results on downstream tasks are achieved by several approaches. BEiT~\cite{ bao2021beit} proposes to recover discrete visual tokens and SimMIM~\cite{Xie2021SimMIMAS} performs pixel-level reconstruction. Recently, MAE \cite{he2022masked}, which recovers pixels from a high masking ratio, has been shown capable of learning robust representations \cite{bachmann2022multimae, convmae, pointm2ae}.
In this work, we focus on designing a mixture strategy to enhance representations learned by MAE~\cite{he2022masked}.

\noindent{\textbf{Image mixtures.}} Widely adopted mixture methods in visual supervised learning include Mixup~\cite{zhang2017mixup} and Cutmix~\cite{yun2019cutmix}. However, these methods require ground-truth labels for calculating mixed labels; in this work, we adapt Mixup to our unsupervised framework by formulating losses on dual reconstructions. 
On the other hand, in recent visual self-supervised learning literature, joint embedding methods and contrastive learning approaches such as MoCo~\cite{he2020momentum}, SimCLR~\cite{chen2020simple}, and more recently UnMix~\cite{shen2022mix} have acquired success and predominance in mixing visual inputs, promoting instance discrimination by aligning features of augmented views of the same image. Related to our work, \cite{mixmim} have proposed to use cutmix in MIM, replacing mask tokens with visible tokens of another image and performing dual reconstructions. Contrarily, we conduct image mixtures at the pixel level rather than token-level, with the unique advantage of being more interpretable with latent decomposition.

\noindent{\textbf{Invariance and disentanglement of representation learning in Autoencoders.}} Representation learning focuses on the properties of features learned by the layers of deep models while remaining agnostic to the particular optimization process. Invariance and disentanglement are two commonly discussed factors that occur in data distribution for representation learning~\cite{DBLP:journals/corr/abs-1912-01991, dvae}. Autoencoders are classical generative unsupervised representation learning frameworks based on image reconstruction as loss function, learning both the mapping of inputs to latent features and the reconstruction of the original input. In this work, we focus on the latent disentanglement where one feature is correlated or connected to other vectors in the latent space in Masked Autoencoder. The motivation for learning disentangled features in Autoencoders is for achieving interpretability~\cite{chen16,bengio} and for intuitive explanations. 

\section{i-MAE} \label{approach}

In this section, we first introduce an overview of our proposed framework. Then, we present our components in detail, including our mixture input, two-branched reconstruction via a shared-decoder, and our patchwise-distillation module. Ensuingly, we elaborate on the metric we propose to evaluate the improved linear separability in i-MAE and the sampling strategy to enhance the degree of semantics in our mixtures, as well as broadly discussing the observations and discoveries.

\subsection{Framework Overview}

\begin{figure}
    \centering
    \hspace{-0.2in}
    \includegraphics[width=0.5\textwidth,height=10cm, keepaspectratio]{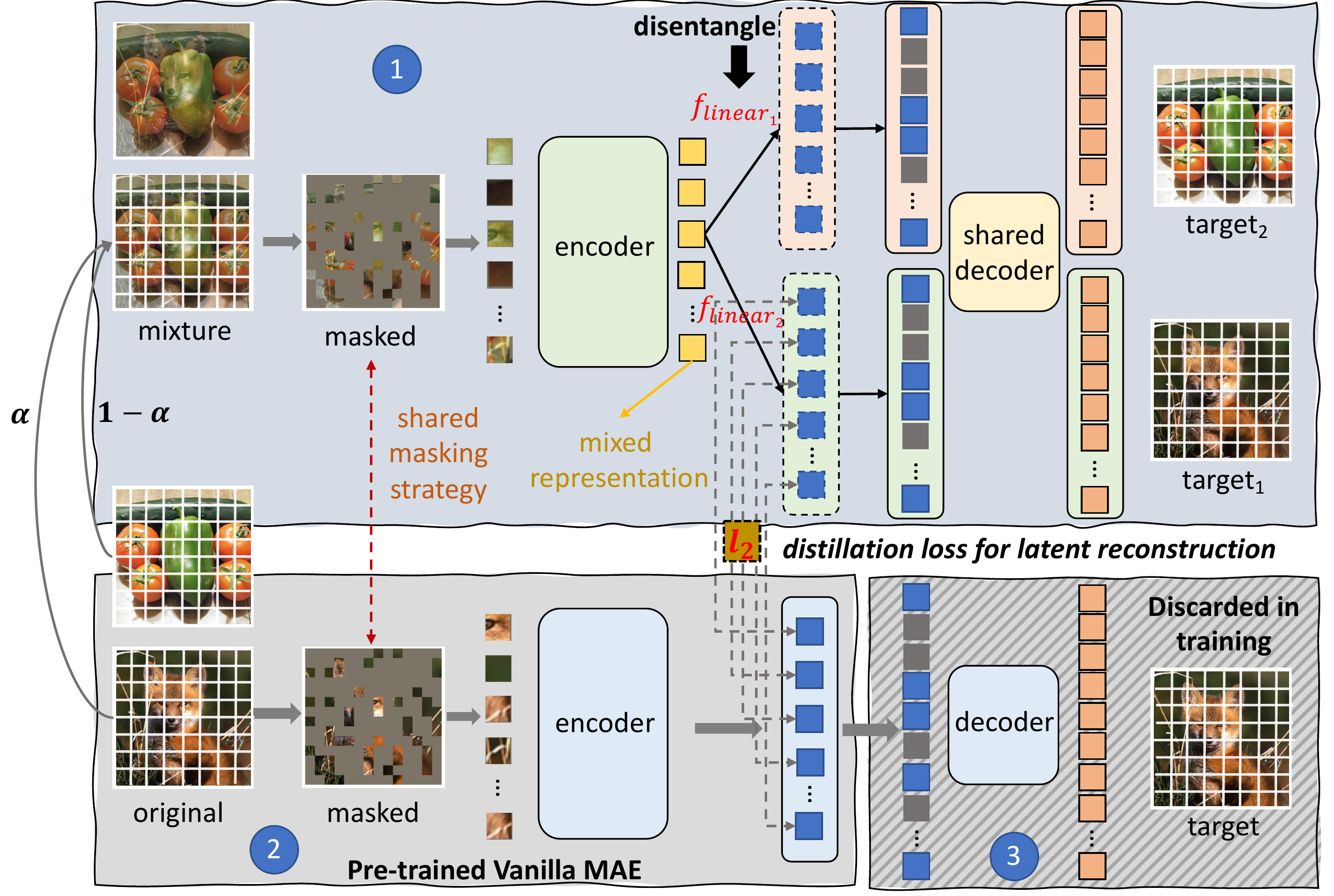}
    \vspace{-0.18in}
    \caption{Framework overview of our {\em i-MAE}. \ding{172} is the main branch that consists of a mixture encoder, a disentanglement module, and a two-way image reconstruction module. \ding{173} is the encoder part of a pre-trained vanilla MAE for distillation purposes (i.e., latent reconstruction). \ding{174} is the decoder part in MAE and is discarded in training.}
    \label{fig:framework}
    \vspace{-0.15in}
\end{figure}

As shown in Fig~\ref{fig:framework}, our framework consists of three submodules: {\bf (1)} a mixture encoder module that takes the masked mixture image as the input and output mixed features; {\bf (2)} a disentanglement module that splits the mixed feature into the individual ones; {\bf (3)} a MAE teacher module that provides the pre-trained embedding for guiding the splitting process in the disentanglement module.

\subsubsection{Components} \label{sec:methods-Two-way}

\textbf{Input Mixture with MAE Encoder.}
Inspired by Mixup~\cite{zhang2017mixup}, we use an unsupervised mixture of inputs formulated by $\bm \alpha *{\textbf I}_1$ and $(1-\bm \alpha)*{\textbf I}_2$, where ${\textbf I}_1, {\textbf I}_2$ are the input images. Essentially, our encoder extrapolates mixed features from a tiny fraction (e.g., 25\%) of visible patches, then we linearly project it to represent both input images separately.  
Formally, the mixed image is:
\begin{equation}
 {\textbf I}_{\bm m} = \bm \alpha * {\textbf I}_1 + (1-\bm \alpha)*{\textbf I}_2,
\end{equation}
where $\bm \alpha$ is the coefficient to mix two images following a Beta distribution. The encoder of i-MAE directly follows MAE~\cite{he2022masked}, generating input tokens from images. The mixed input is divided into non-overlapping patches and then goes through the embedding procedure. The same masking strategy is used in both the teacher (Vanilla MAE) and student (i-MAE) models. 

\noindent{\textbf{Two-branch Masked Autoencoders with Shared Decoder.}} 
Although sufficient semantic information from both images is embedded in the mixed representation to reconstruct both images, the vanilla MAE cannot by itself associate the entangled features with either input. The MAE structure does not retain identification information (e.g., order or positional information) about the two inputs that are mixed in image space, i.e., the model cannot tell which of the two images to reconstruct to, since both are sampled from the same distribution and mixed randomly. The consequence is that both reconstructions look identical to each other and fail to look similar to either original input.

Similar to how positional embeddings are needed to explicitly encode spatial information, i-MAE implicitly encodes the semantic difference between the two inputs by using a dominant and subordinate mixture strategy. In practice, we train the linear separation layers to distinguish between the dominant input $\textbf I_d$ (higher mix factor) and the subordinate input $\textbf I_s$ (lower mix factor). 

\noindent{\textbf{Two-way Image Reconstruction Loss.}}
Formally, we build our reconstruction loss to recover individual images from a mixed input, which is first fed into the encoder to generate mixed features:
\begin{equation}
   \textbf h_m = \textbf{E}_\text{i-MAE}(\textbf I_m),
\end{equation}
where $\bm E_\text{i-MAE}$ is i-MAE's encoder, $\textbf h_m$ is the latent mixed representation. Then, we employ two non-shareable linear embedding layers to separate the mixed representation into individual ones:
\begin{equation}
  \begin{aligned} \textbf h_1 = \mathcal{F}_1(\textbf h_m), \\\textbf h_2 = \mathcal{F}_2(\textbf h_m), \end{aligned}
\end{equation}
where $\mathcal{F}_1$, $\mathcal{F}_2$ are two linear layers with different parameters for disentanglement, and $\textbf h_1$ and $\textbf h_2$ are corresponding representations. After that, we feed the individual representations into the shared decoder with the corresponding reconstruction losses:
\begin{equation}
\begin{array}{l}\mathcal{L}_{\text {recon}}^{\textbf{I}_{1}}=\mathbb{E}_{\textbf{I}_{1} \sim p\left(\textbf{I}_{1}\right)}[\left\|\textbf{D}_{\text {shared}}\left(\textbf{h}_{1}\right)-\textbf{I}_{1}\right\|_{2}], \\ \mathcal{L}_{\text {recon}}^{\textbf{I}_{2}}=\mathbb{E}_{\textbf{I}_{2} \sim p\left(\textbf{I}_{2}\right)}[\left\|\textbf{D}_{\text {shared}}\left(\textbf{h}_{2}\right)-\textbf{I}_{2}\right\|_{2}],\end{array}
\end{equation}
where $p(I_n)$ is the sample distribution of input images.

We have shown that our encoder learns to embed representations of both images. We also examined reconstructing only the subordinate image $\textbf I_s$ to prevent the $\textbf I_d$ from guiding the reconstruction. We empirically verified that a single branch is sufficient for the proposed framework to converge training and reconstruct the subordinate target. We provide comparisons with the representational abilities of the single branch i-MAE in the experiments. By default, we choose the double-branched i-MAE for stronger representation performance. 
Essentially, successful reconstructions from only the $\textbf I_s$ prove that representations of both images can be learned and that the subordinate image is not filtered out as noise.

Concretely, through an unbalanced mix ratio and a reconstruction loss targeting only one of the inputs, our framework encodes sufficient information for i-MAE to linearly map the input mixture to two outputs.

\noindent{\textbf{Patch-wise Distillation Loss for Latent Reconstruction.}}
With the linear separation layers and an in-balanced mixture, the i-MAE encoder is presented with sufficient information about both images to perform visual reconstructions. However, information is inevitably lost during the mixing process, harming the value of the learned features in downstream tasks such as classification. To mitigate such an effect, we propose a knowledge distillation module for not only enhancing the learned features' quality, but also demonstrating that a successful distillation can evidently prove the linear separability of our features. 

Intuitively, MAE's features can be regarded as ``ground-truth'' and i-MAE learns features distilled from the original MAE. Specifically, our loss function computes $\ell_2$ loss between disentangled representations and original representations to help our encoder learn useful features of both inputs. 
Our Patch-wise latent reconstruction loss can be formulated as:
\begin{equation}
\begin{aligned} \mathcal{L}_{\text {recon }}^{\textbf h_{1}} &=\mathbb{E}_{\textbf{h}_{1} \sim q\left(\textbf{h}_{1}\right)}[\left\|\textbf{E}_{\mathrm{p}-\mathrm{MAE}}\left(\textbf{I}_{1}\right)-\textbf{h}_{1}\right\|_{2}], \\ \mathcal{L}_{\text {recon }}^{\textbf{h}_{2}} &=\mathbb{E}_{\textbf{h}_{2} \sim q\left(\textbf{h}_{2}\right)}[\left\|\textbf{E}_{\mathrm{p}-\mathrm{MAE}}\left(\textbf{I}_{2}\right)-\textbf{h}_{2}\right\|_{2}],\end{aligned}
\end{equation}
where $\bm E_\text{p-MAE}$ is the pre-trained MAE encoder.

\noindent{\textbf{Semantics-enhanced Sampling.}} 
Our {\em semantics-enhanced sampling} is a data sampling strategy that introduces significantly more image mixtures belonging to the same class into the training process, which boosts the semantics learned by the double-branched i-MAE.
Specifically, we select training instances from the same classes following different distributions to constitute an input mixture as follows: 
\begin{equation}
    \bm p = \mathcal{F}_{\bm m}(\textbf I_{c_a}+\textbf I_{c_b}),
\end{equation}
where ${\mathcal{F}}_{\bm{m}}$ is the backbone network for mixture input and ${\bm p}$ is the corresponding prediction. ${\textbf I}_{c_a}$ and ${\textbf I}_{c_b}$ are the input samples, and $c_a$, $c_b$ have a certain percentage ${\bm r}$ that belongs to the same category. For instance, if ${\bm r}=0.1$, it indicates that 10\% images in a mini-batch are mixed with the same class. When ${{\bm r}}=1.0$, all training images will be mixed with another one from the same class. 
We train the model with a fixed ${\bm r}$, and find that the intuitive method can effectively enhance learned semantics. 

\subsection{Linear Separability} \label{linear_method}
In this section, we explain the approaches we utilize to measure linear separability of the features learned in i-MAE. Many classical works on Autoencoders have demonstrated the improved interpretability from disentanglement~\cite{bengio} and especially linear disentanglement~\cite{LinearSep, LinearDisentangle}. In i-MAE, the motivation behind our pre-training strategy is for more linearly separable features.

For i-MAE to reconstruct both the subordinate and dominant image from a potentially linear mixture, 
not only should the encoder be general enough to retain information from both inputs, but it also needs to generate embeddings that are specific enough for the decoder to distinguish them into their pixel-level forms. A straightforward interpretation of how i-MAE fulfills both conditions is that the latent mixture ${{\textbf h}_m}$ is a linear combination of features that closely relate to ${\textbf h_1}$ and ${\textbf h_2}$, e.g., in a linear relationship. 
To verify this explanation, we employ a linear separability metric to experimentally observe such a behavior.

\noindent{\textbf{Metric of Linear Separability.}}  
\label{sec:method_lin_sep}
A core contribution of our i-MAE is that the framework learns features that are better linearly disentangled, and we provide tools to evaluate this observation. In general, linear separability is a property of two sets of features that can be separated into their respective sets by a hyperplane. In our example, the set of latent representations $\textbf H_1$ and $\textbf H_2 $ are linearly separable if there exists $ n+1$ real numbers $\textbf w_1,\textbf w_2, ...,\textbf w_n,\textbf b$, such that every $\textbf h \in \textbf H_1$ satisfies $\sum{{\textbf w}_i{\textbf h}_i > \textbf b}$ and every $\textbf h \in \textbf H_2$ satisfies $\sum{{\textbf w}_i{\textbf h}_i < \textbf b}$. It is a common practice to train a linear classifier (e.g., SVM~\cite{stylegan}) or a linear regressor~\cite{Eastwood2018AFF, 10.5555/3104482.3104547} for evaluating if the two sets of features are linearly separable. 

{To measure the linear relation of disentanglement, we train a linear regressor with with $\ell_1$ regularization (lasso penalty) between the disentangled features of the subordinate image $\textbf I_s$ and the vanilla MAE features of the same input $\textbf I_s$.} Intuitively, since the disentangled features without constraints will be far from the features of the vanilla MAE model, we utilize a linear regressor to fit the disentangled features to the vanilla features for comparisons. 
To quantitatively measure the linear separability of i-MAE, we utilize a variety of scores and distances to evaluate the manifold correlations, including {\em Normalized Root Mean Square Error (NRMSE)}, {\em Coefficient of Determination (R-squared $R^2$)}, and {\em Cosine Similarity}. Among them, NRMSE and R-squared are used to evaluate the correlation of regression, and cosine distance can measure how close the disentangled features are with the original features when mapped to the same latent space.

\subsection{Semantics} \label{sema_method}
\noindent{\textbf{Enhanced Semantics.}} 
In our mixture strategy and double reconstructions, the representational ability learned is affected by the mixing strategy; hence, we improve our learning framework with semantics-enhanced sampling, an approach that samples image mixtures pertaining to the same class. Intuitively, the disentanglement of features from the same class is more difficult than segmenting different classes; intra-class separation necessitates knowledge of high-level visual concepts, such as semantic differences, rather than lower-level patterns, such as shape or color. 
Moreover, when mixing images of the same class, latent features are naturally more similar, and their two-way loss functions will be updated in the same direction. Consequently, an intra-class mixture's latent features will encode more information that is more robust about a specific class than an inter-class mixture, where the two latent features may confuse or conflict with each other. Specifically, when the mixed representations have semantics that are more closely aligned, the information propagated into the two branches contains more information about the same class. Otherwise, when the mixed representations are from different classes, the disentangled features may not have semantics perfectly resembling their classes.

From these two observations, we introduce {\em semantics-enhanced sampling} for generating different percentages of same-classed mixtures ${\bm r}$, a hyper-parameter we experimentally verify. After the model is trained by i-MAE using such kind of input data, we finetune the model with Mixup strategy (both baseline and our models) and cross-entropy loss. We use accuracy as the metric of semantics under this percentage of instance mixture:
\begin{equation}
    \mathcal M_\text{sem} =-\sum_{i=1}^{n} \textbf t_{i} \log \left(\textbf p_{i}\right),
\end{equation}
where $\textbf t_i$ is the ground-truth label and $\textbf p_{i}$ is the prediction.
Since this sampling strategy will involve additional prior knowledge of same or different classes when sampling and mixing in a mini-batch, an alternative way to avoid using prior label information is clustering the samples to identify the same or different classes in a mini-batch, then conducting the sample assignment process.

\section{Experiments and Analysis}

In this section, we examine the effectiveness of the proposed i-MAE framework and analyze the properties of i-MAE's disentangled representations through empirical studies on an extensive range of datasets. First, we provide the details of datasets used and our implementation settings. Then, we thoroughly ablate our experiments, focusing on the linear and finetuning evaluations, as well as properties of { linear separation}, and { semantic-enhanced mixture}. Lastly, we illustrate the qualitative results and visualizations.

\subsection{Datasets}
\textbf{CIFAR-10/100}~\cite{Krizhevsky2009LearningML} Both CIFAR datasets contain 60,000 tiny colored images sized 32$\times$32. CIFAR-10 and 100 are split into 10 and 100 classes, respectively. 

\noindent{\bf Tiny-ImageNet} The Tiny-ImageNet is a scaled-down version of the standard ImageNet-1K consisting of 100,000 64x64 colored images, categorized into 200 classes.

\noindent{\bf ImageNet-1K}~\cite{deng2009ImageNet} The ILSVRC 2012 ImageNet-1K classification dataset consists of 1.28 million training images and 50,000 validation images of 1000 classes.

\subsection{Details of Implementation} 
\textbf{Settings}: We conduct experiments of i-MAE on CIFAR-10/100, Tiny-ImageNet, and ImageNet-1K. On CIFAR-10/100, we adjust MAE's structure to better fit the smaller datasets during unsupervised pre-training: ViT-Tiny~\cite{deit} in the encoder and a lite-version of ViT-Tiny (4 layers) as the decoder. Our pre-training lasts 2,000 epochs with a learning rate $1.5\times10^{-4}$ and 200 warm-up epochs. On Tiny-ImageNet, i-MAE's encoder is ViT-small and decoder is ViT-Tiny, trained for 1,000 epochs with a learning rate $1.5\times10^{-4}$. Additionally, we apply warm-up for the first 100 epochs, and use cosine learning rate decay with AdamW~\cite{adamw} optimizer as in vanilla MAE. 

Unless otherwise stated, the default settings used in our experiments are a masking ratio of 75\%, a mix factor sampled from a distribution $\beta(1.0, 1.0)$, and reconstructing both images with distillation loss for stronger representation. 

\noindent{\bf Supervised Finetuning}: 
In the finetuning process, we apply Mixup in all experiments to fit our pre-training scheme, and compare our results with baselines of the same configuration. On CIFAR-10/100, we finetune 100 epochs using the AdamW optimizer and a learning rate of $1.5\times10^{-3}$.

\noindent{\bf Linear Probing}: 
For linear evaluation, we follow MAE~\cite{he2022masked} to train with no extra augmentations and use zero weight decay. Similarly, we adopt an additional BatchNorm layer without affine transformation. 

\subsection{Main Results}

In this section, we first provide the finetuning and linear evaluation results on various datasets. Following that, we empirically analyze our main findings: how separable are i-MAE embedded features and the advantage of semantics-enhanced sampling. Specifically, we quantitatively verify the linear separability of i-MAE's disentanglement. Then, we evaluate the performance improvement from utilizing the semantics-enhanced sampling strategy. 

\noindent{\textbf{Finetuning and Linear Evaluation.}} 
Here we explore the performance gain from the architecture and training strategy adjustments. 
We evaluate our i-MAE's performance through finetuning and linear evaluation of regular inputs and targets. 
Finetuning and linear probing classification results are outlined in Tab.~\ref{fig:class_acc_finetune} and Tab.~\ref{fig:class_linear_acc}. It can be observed that i-MAE outperforms the baseline MAE and ViT from scratch by remarkable margins. As our features are learned from a harder scenario, they encode more information with more robust representation and classification accuracy. Besides, i-MAE shows a considerable performance boost with both evaluation methods. 

\begin{figure}[t]
    \centering
    \includegraphics[width=0.95\linewidth, height=12cm, keepaspectratio]{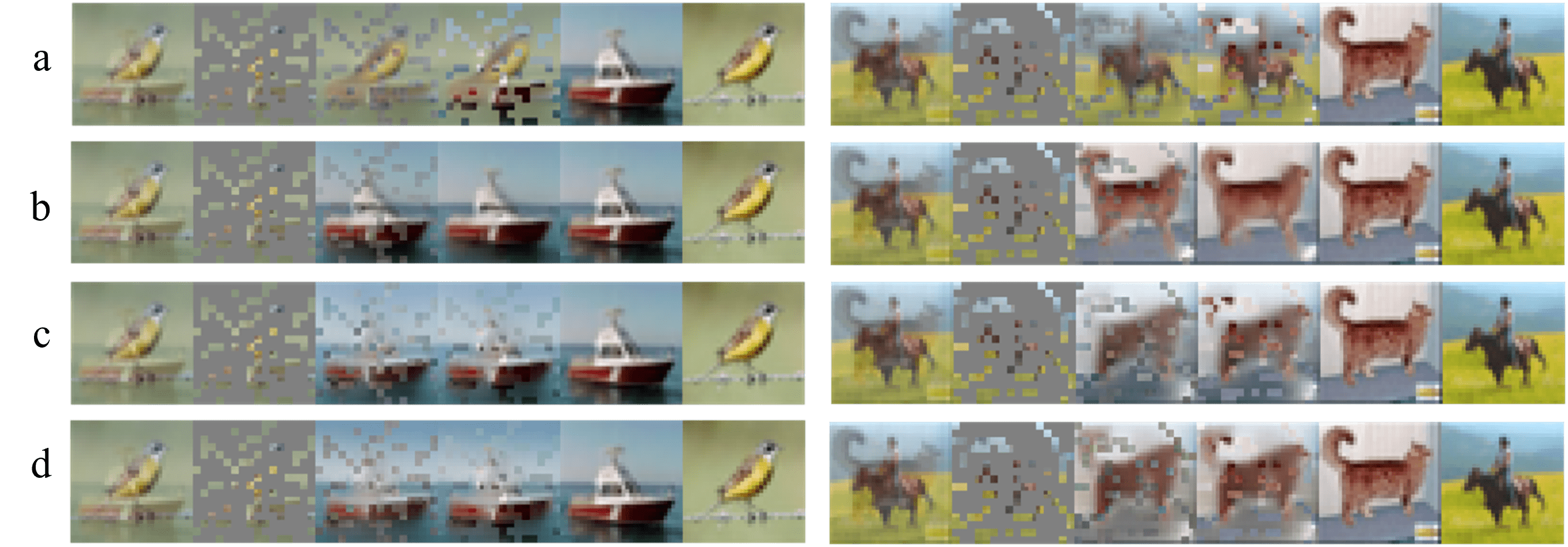}
    \vspace{-0.10in}
    \caption{Qualitative ablation comparisons on CIFAR-10. {\bf Row (a)}: baseline vanilla MAE; {\bf (b)}: MAE with unmixed input; {\bf (c)}: our i-MAE without distillation; and {\bf (d)}: i-MAE with distillation.} 
    \label{fig:visual comparison}
    \vspace{-0.11in}
\end{figure}

\noindent{\textbf{Separability.}} 
Here, we show how i-MAE displays properties of linear separability, both visually and quantitatively, and demonstrate our advantage over vanilla MAE. 
We provide a visual comparison of the disentanglement capability in Fig.~\ref{fig:visual comparison}. 
In the first row, vanilla MAE does not perform well out-of-the-box when disentangling mixed inputs, with reconstructions representing the mixed input more so than the subordinate image. However, the latter two rows demonstrate that i-MAE performs reconstruction very well.

Since the mixture inputs of i-MAE is a linear combination of the two images, and our results show i-MAE's powerful ability to reconstruct both images, even at very low mixture ratios, we attribute such ability to i-MAE's disentanglement strongly correlating with the vanilla MAE's features.  
Now, we empirically illustrate the strength of the linear relationship between MAE's features and i-MAE's disentangled features with a linear separability metric. We employ different distances as our criteria, and results are reported in Tab.~\ref{tab:linear_separation}. Experimentally, we feed mixed inputs to i-MAE and a singular image to vanilla MAE, where the former produces disentangled features and the latter produces target features. Then, we train a single linear projection layer to fit the disentangled features to the target.
{\em Fore} indicates that we directly calculate the metrics between the target features from vanilla MAE and our disentangled features. {\em Aft} indicates that we train the linear regressor's parameters to fit the target feature.  {\em Baseline} is the model trained without the disentanglement module. It can be observed that our i-MAE has significantly smaller distances than the vanilla model, indicating that such a scheme can obtain better linear separation ability.

\begin{table}[t]
\centering
 \resizebox{0.44\textwidth}{!}{
\begin{tabular}{l | c| c | c  } 

 Method & CIFAR-10 & CIFAR-100 & Tiny-ImageNet  \\ \hline
from scratch & 74.13 &  53.57 & 43.36 \\ 
 MAE & 90.78 & 68.66 & 59.28  \\ 
 \bf i-MAE &  \bf 92.00 & \bf 69.50 & \bf 61.63 \\  

\end{tabular}
}
\vspace{-0.05in}
\caption{Finetuning classification accuracy of ViT trained from scratch, baseline MAE, and the proposed i-MAE across different datasets. }
\label{fig:class_acc_finetune}

\resizebox{0.44\textwidth}{!}{
\begin{tabular}{l | c| c | c } 

 Method & CIFAR-10 & CIFAR-100 & Tiny-ImageNet  \\ \hline
 MAE &  72.47 &  32.57 &  19.62  \\ 
 \bf i-MAE & \bf 77.61 & \bf 33.39 & \bf 20.40  \\   
\end{tabular}
}
\vspace{-0.05in}
\caption{Linear probing accuracy of baseline MAE and proposed i-MAE across different datasets. {\em from scratch} is inapplicable here.}
\label{fig:class_linear_acc}
\vspace{-0.1in}
\end{table}

\begin{table*}[t]
    \begin{center}
     \resizebox{0.6\textwidth}{!}{
    \begin{tabular}{c | c  c | c c| cc} 
     \midrule
 \multirow{2}{*}{Same-class Ratio}  & \multicolumn{2}{| c |}{CIFAR-10} & \multicolumn{2}{| c }{CIFAR-100} & \multicolumn{2}{| c }{Tiny-ImageNet} \\
     \cmidrule(r){2-7}
      & Finetune & Linear & Finetune & Linear & Finetune & Linear\\ [0.5ex] 
    \midrule
     MAE (baseline) & 90.78 & 72.47 & 68.66 &  32.57 & 59.28& 19.62\\ 
     \midrule
     0.0 & 91.67 & 70.53 & 68.34 & 29.22 &  60.91 & 18.23\\ 
     0.5 & \bf 92.34 & 72.80 & \bf 69.50 & 30.11 & 60.58 & 18.51 \\
     1.0 & 91.60 & \bf 77.61 &  69.33 & \bf 33.39 & \bf 61.13 & \bf 20.40 \\  
     \midrule
    \end{tabular}
    }
    \vspace{-0.1in}
        \caption{Ablation study of semantics-enhanced sampling strategy with intra-class mix rate $\bm r$ from 0.0 to 1.0 ({\em baseline} is the vanilla MAE, 0.0 indicates the regular sampling strategy). The lower bound represents that inputs are mixes with all different classes, $\bm r =1.0$ indicates the model is pre-trained with solely mixtures of same-labeled instances. }
    \label{tab:ft_sem}
    \end{center}
    \vspace{-0.15in}
\end{table*}

\begin{table*}[t]
\begin{center}
 \resizebox{0.75\textwidth}{!}{
\begin{tabular}{ c |  c | c c |  c c | c c | c c} 
  \midrule
  \multirow{2}{*}{Score} & \multirow{2}{*}{Method} & \multicolumn{2}{ c |}{CIFAR-10} & \multicolumn{2}{| c| }{CIFAR-100} & \multicolumn{2}{| c |}{Tiny-ImageNet} & \multicolumn{2}{| c }{ImageNet-1K} \\
    & & Fore & Aft  & Fore & Aft & Fore & Aft  & Fore & Aft  \\  
 \midrule
 \multirow{3}{*}{NRMSE  $\downarrow$} &
 MAE (Baseline)            & 0.247  &  0.223  & 0.254  & 0.208 & \bf 0.701 &   0.624  \bf & \bf 0.639  & 0.334 \\ 

 & i-MAE w/o distill & 0.372  &  0.221 &   0.368 &  0.202  & 1.023 &  0.627 & 0.850  &  0.313 \\ 

 & \bf i-MAE             & \bf 0.179 & \bf 0.188 & \bf 0.176 & \bf 0.181 & 0.927 & \bf 0.597 &  0.869  &   \bf 0.299 \\ % 
 \midrule

  \multirow{3}{*}{$R^2$ $\uparrow$}  &
 MAE (Baseline)            & 0.336  &  0.438  & 0.242  & 0.482 & 0.269 &   0.429   & -  & 0.625 \\ 

 & i-MAE w/o distill & -  &  0.540 & -  & 0.513 & - &  0.420  & -  &  0.714\\ 

 & \bf i-MAE             & \bf 0.807 & \bf 0.637 & \bf 0.633 & \bf 0.609 & \bf 0.275 & \bf 0.480 & -  & \bf   0.736 \\ 
 \midrule

  \multirow{3}{*}{Cos $\uparrow$}  &
 MAE (Baseline)            & 0.672  &  0.665   & 0.626  & 0.694 & \bf 0.643 &  0.662   & \bf 0.501  & 0.768 \\ 

 & i-MAE w/o distill & 0.062  &  0.741 &  0.007  &  0.718 & -  &  0.655  & 0.002  &  0.816  \\ 

 & \bf i-MAE             & \bf 0.807 & \bf 0.796 & \bf 0.794 & \bf 0.779 &  0.000 & \bf 0.696 &  0.009  & \bf 0.837 \\ % 
 \midrule
\end{tabular}
}
\vspace{-0.13in}
\caption{Linear separation metric using NRMSE, $R^2$ (R-Squared or coefficient of determination) and {\em cosine}-similarity as measurements for patch-wise feature comparisons, calculated before and after linear regression on CIFAR-10, CIFAR-100, Tiny-ImageNet, and ImageNet. Reported results are from linear regressor's prediction on validation set. 
{\em i-MAE} and {\em i-MAE without distillation} are embedding after disentanglement. $\downarrow$ indicates lower is better and $\uparrow$ indicates higher is better.
}
\label{tab:linear_separation}

\end{center}
\vspace{-0.15in}
\end{table*}

\noindent{\textbf{Semantics.}} As shown in Tab.~\ref{tab:ft_sem}, we provide the ablation study of semantics-enhanced sampling strategy with intra-class mix rate r from 0.0 to 1.0. In the table, {\em baseline} represents the vanilla MAE model and 0 is the regular sampling strategy without semantics-enhanced sampling. We can observe that the performacne has a certain increase when the same-class ratio is employed with 0.5 or 1.0.

As discussed in the Sec. \ref{sec:method_lin_sep}, we emphasize that our enhanced performance comes from i-MAE's ability to learn more separable features with the disentanglement module, and the enhanced semantics learned from training with {\em semantics-enhanced sampling}. Our classification results show the cruciality of MAE learning features that are linearly separable, which can help identify between different classes. However, to correctly identify features with their corresponding classes, semantically rich features are needed, which can be enhanced by the intra-class mix sampling strategy.

\subsection{Ablation Study}
In this section, we perform ablation studies on i-MAE to concretely examine the property of linear separability and its existence at different mix-levels. Then, we analyze the effects of {\em semantics-enhanced mixture} on i-MAE learned representations.

\noindent{\textbf{Ablation for Linear Separability.}} 
In this work, motivated by approaches measuring latent disentanglement with linear means, we propose mixture strategies and semantic mixing for the purpose of learning stronger and more linearly separable features.
To begin, we thoroughly perform our ablation experiments on a diverse group of datasets (ImageNet-1K is performed for final evaluation) and demonstrate how i-MAE's learned features display linear separability with different settings. Specifically, we experiment with the separability of the following aspects of our methods: (1) constant and probability mix factors; 
(2) masking ratio of input mixtures; (3) different ViT architectures. 
As previously mentioned, since our model produces two reconstructions, our visual results demonstrate the subordinate branch, which is worse in quality, to demonstrate the effectiveness of our method.

\begin{figure}
    \centering
    \includegraphics[width=0.99\linewidth, height=4cm, keepaspectratio]{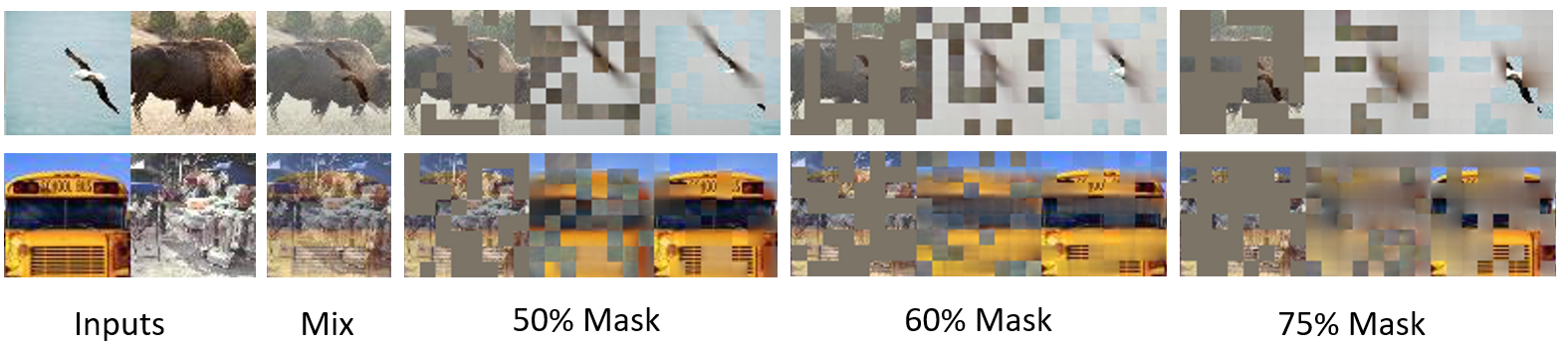}
    \vspace{-0.15in}
    \caption{Comparisons between mask ratios on Tiny-ImageNet validation set. i-MAE produces enhanced visual reconstructions from lower masking ratios when reconstructing images. }
    \label{fig:maskratio}
    \vspace{-0.1in}
\end{figure}

\noindent{\textbf{(1) Mix Ratio.}} 
To demonstrate the separable nature of the input mixtures for reconstructions, we compared different mixture factors between 0 and 0.5, and random mixture ratios from a Beta distribution. Intuitively, lower mixing ratios contain less meaning information that the encoder may easily confuse with noise, whereas higher ratios destroy the subordinate-dominant relationship. Experimentally, we observe matching visual results shown in the supplementary materials The better separation performance near 0.3 indicates that i-MAE features are better dichotomized when the mix factor is balanced between noise and useful signals. Whereas below 0.15, the subordinate image is noisy and reconstructions are not interpretable, mixing ratios above 0.45 break the subordinate relationship between the two images, and the two features are harder to distinguish from each other. Moreover, Fig.~\ref{fig:ImageNet} presents a problematic case where a mix factor of 0.45 reconstructs dominant features (hence the green patches in the background).

\noindent{\textbf{(2) Mask Ratio.}} 
In i-MAE, visible information of the subordinate image is inherently limited by the unbalanced mix ratio, in addition to masking. Hence, a high masking ratio (75\%~\cite{he2022masked}) may not be necessary to suppress the amount of information the encoder sees, and we consequently attempt lower ratios of 50\%, 60\% to introduce more information about the subordinate target. As shown in Fig.~\ref{fig:maskratio}, a lower masking ratio of 0.5 or 0.6 can significantly improve reconstruction quality.

Combining our findings in mix and mask ratios, we empirically find that i-MAE can compensate for the information loss at low ratios with the additional alleviation of more visible patches (lower mask ratio). Illustrated in Fig.~\ref{fig:ImageNet}, we display a case of i-MAE's reconstruction succeeding in separating the features an input with {\bf $\bm \alpha=0.1$} mix factor and {0.5} masking ratio. Through studying the mix ratio and masking ratio, we reveal that i-MAE can learn linearly separable features under two conditions: {\bf (i)} enough information about both images must be present (determined by the trade-off between mask ratio and mix ratio). {\bf (ii)} the image-level distinguishing relationship between minority and majority (determined by mix ratio) is potent enough for i-MAE to encode the two images separately.

\noindent{\textbf{(3) ViT Backbone Architecture.}} 
We studied whether different scales of ViT effect linear separation in Appendix. Our results show that larger backbones are not necessary for i-MAE to disentangle features on small datasets, as the insufficient training data cannot fully utilize the capability. However, large ViTs are crucial to the large-scale ImageNet-1K dataset.

\noindent{\textbf{Ablation for Degree of Semantics.}} 
We provide the ablation study on different ratios of mixing the same class samples within a mini-batch.

\noindent{{\em Semantic Mixes.}}  
Depending on the number of classes and their overall size, datasets in pristine states usually contain around 10\% (e.g., CIFAR-10) to $<$1\% (e.g. ImageNet-1K) samples pertaining to the same class, meaning that by default, uniformly random sampling mixtures will most likely be of different objects. On the other hand, the {\em semantics-enhanced mixture} scheme examines whether the introduction of semantically homogeneous mixtures affects the classification performance. That is, we intentionally test to see if similar instances during pre-training negatively influence the classification performance. 

As shown in Tab.~\ref{tab:ft_sem}, after i-MAE pre-training, we perform finetuning and linear probing on classification tasks to evaluate the degree of semantics learned given different amounts of intra-class mix $\bm r$. From Tab.~\ref{tab:ft_sem}, we discover that i-MAE overall has a stronger performance in finetuning and linear probing with a non-zero same-class ratio. Specifically, a high $\bm r$ of 1.0 increases the accuracy in linear evaluation most in all datasets, meaning that the quality of learned features is best and separated, and it gains a strong prior of category information for semantically enhanced mixtures. On the other hand, setting $\bm r=0.5$ is advantageous during finetuning, as it gains a balanced prior of separating both intra- and inter-class mixtures. 

\begin{table}[t]
\centering
 \resizebox{0.44\textwidth}{!}{
\begin{tabular}{c | c| c | c  } 
 Method & CIFAR-10 & CIFAR-100 & Tiny-ImageNet  \\ 
 \hline
 i-MAE-Sub & 74.23 & 49.83 & 49.50  \\ 
 \bf i-MAE &  \bf 92.00 & \bf 69.50 & \bf 61.63 \\   
\end{tabular}
}
\vspace{-0.1in}
\caption{Single and dual reconstruction ablation. We compare i-MAE pre-trained to only reconstruct the subordinate image (i-MAE-sub) and to reconstruct both (i-MAE). Better representations are learned with dual reconstructions. }
\label{fig:class_acc_one_two}

\vspace{-0.12in}
\end{table}

\noindent{\textbf{Single vs. Dual Reconstruction}} 

To verify the design in Sec.~\ref{sec:methods-Two-way}, we perform finetuning with i-MAE pre-trained on linear disentanglement with single and dual reconstructions. The results are shown in Table~\ref{fig:class_acc_one_two}. It is clear that dual reconstructions achieve better representational performance.

\subsection{Visualizations}
\begin{figure}
    \centering
    \includegraphics[width=0.48\linewidth, keepaspectratio]{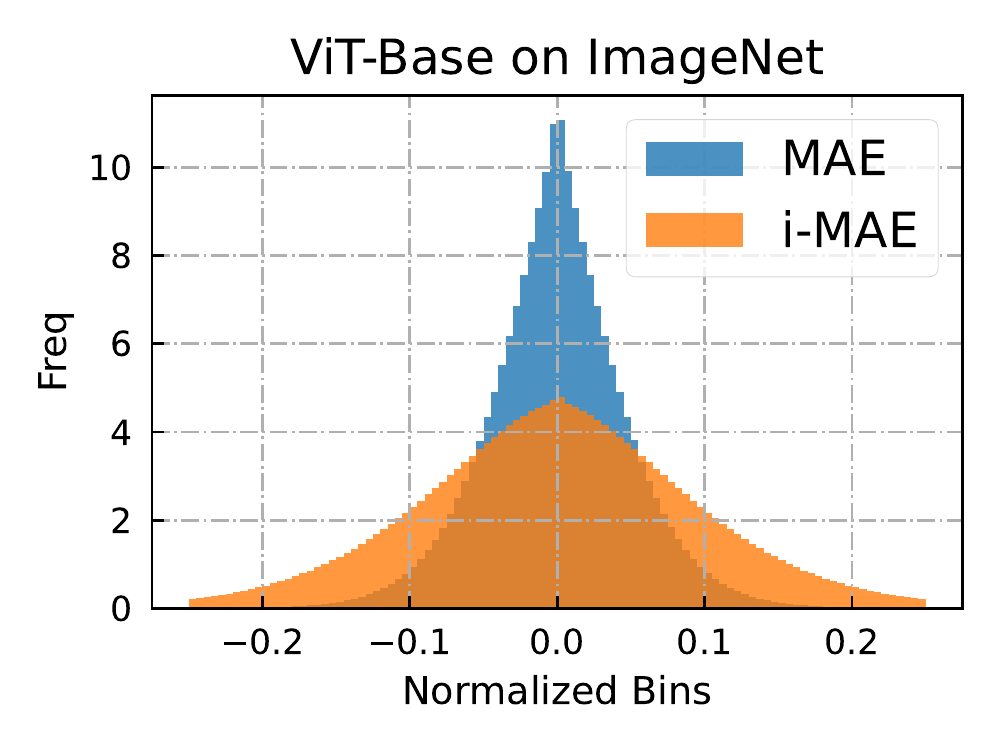}
    \includegraphics[width=0.4\linewidth, keepaspectratio]{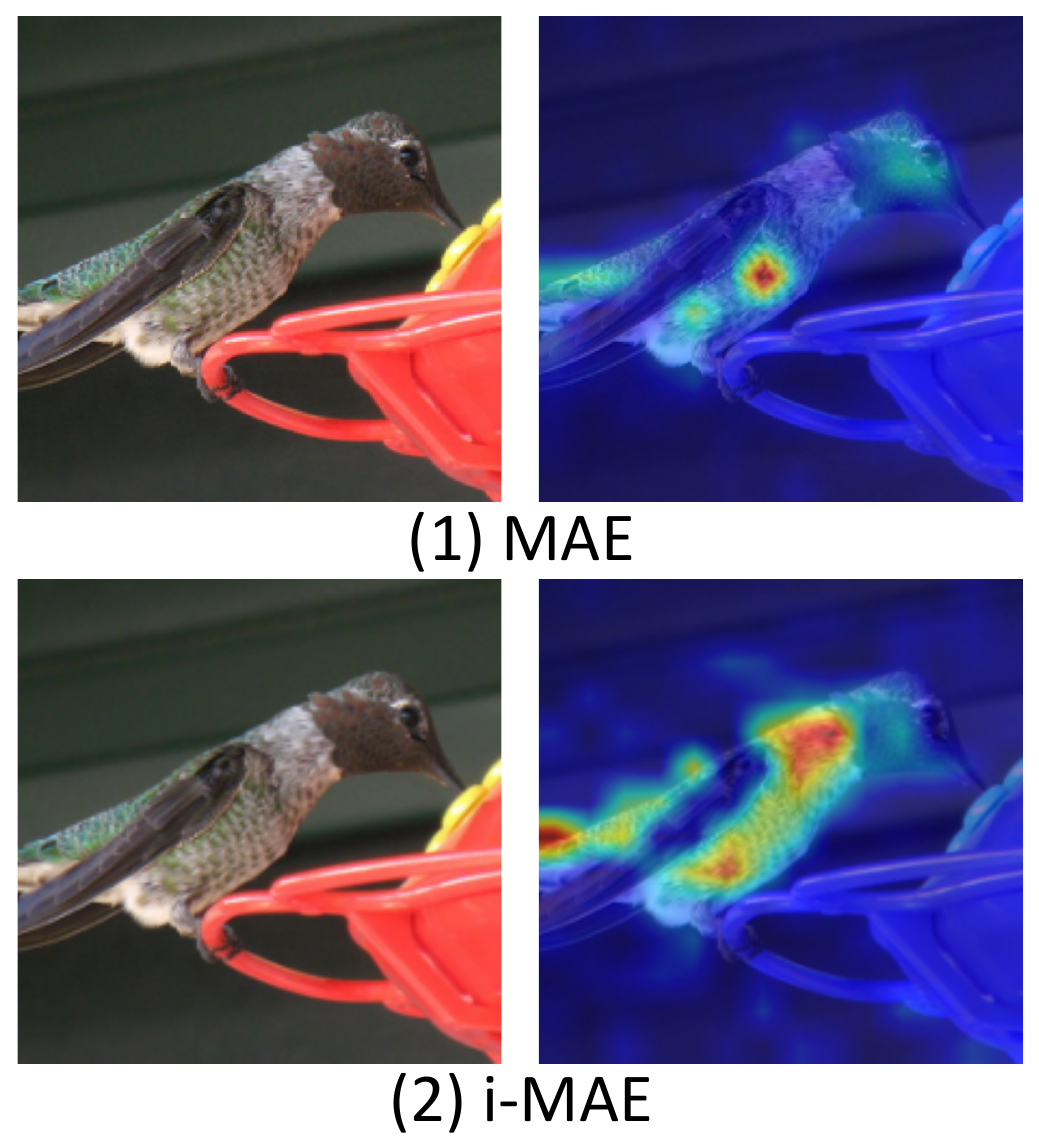}
    \vspace{-0.12in}
    \caption{Left is the comparison of weight distribution between MAE and i-MAE pre-trained on ImageNet-1K. Our weights are sparser. Right is the comparison of attention maps. Our model has a wider view on the input image which encodes more information.}
    \label{fig:weight_dis}
    \vspace{-0.12in}
\end{figure}

We provide the visualization comparison of weight distributions and attention mappings. In Fig.~\ref{fig:weight_dis} (left), we show the difference of weight distributions between MAE and i-MAE on ImageNet-1K dataset. Compared with vanilla MAE's weights, our model's weights have a better diversification distribution, indicating that i-MAE forces the model to spread out and incorporate more patterns over the entire image mix range. In Fig.~\ref{fig:weight_dis} (right), our model has a wider focusing area, which also indicates that more information is encoded in the trained model through the proposed mixture training scheme.

\section{Conclusion}

It is nontrivial to understand why Masked Image Modeling (MIM) in the self-supervised scheme can learn useful representations for downstream tasks without labels. In this work, we have introduced a novel interactive framework upon Masked Autoencoders (i-MAE) to explore two critical properties in latent features: {\em linear separability} and {\em degree of semantics}. We identified that the two specialties are the core for superior latent representations and revealed the reasons where is the good transferability of MAE from. Moreover, we proposed two metrics to evaluate these two specialties quantitatively. Extensive experiments are conducted on CIFAR-10/100, Tiny-ImageNet, and ImageNet-1K datasets to demonstrate our discoveries and observations in this work. We also provided sufficient qualitative results and analyses of different hyperparameters. We hope this work can inspire more studies on the understanding and improvement of the MIM frameworks for the self-supervised pretraining in the future.

{
    \small
    \bibliographystyle{ieeenat_fullname}
    \bibliography{main}
}

\clearpage
\appendix

\section*{Appendix}
In the appendix, we provide the detailed configurations of our experiments and elaborate with more visualizations that are supplementary for our main text, specifically: 

•  Section~\ref{details}  ``Implementation Details'': implementation details and configuration settings for unsupervised pre-training and supervised classification. 

•  Section~\ref{vis_app} ``More Visualizations'': additional reconstruction examples on various datasets.

•  Section~\ref{pseu} ``Pseudocode'': a PyTorch-like pseudocode for our detailed procedure of the proposed framework.

\section{Implementation Details in Self-supervised Pre-training, Finetuning, and Linear Evaluation} \label{details}

\begin{figure}[h]
    \centering
    \includegraphics[width=0.96\linewidth, height=20cm, keepaspectratio]{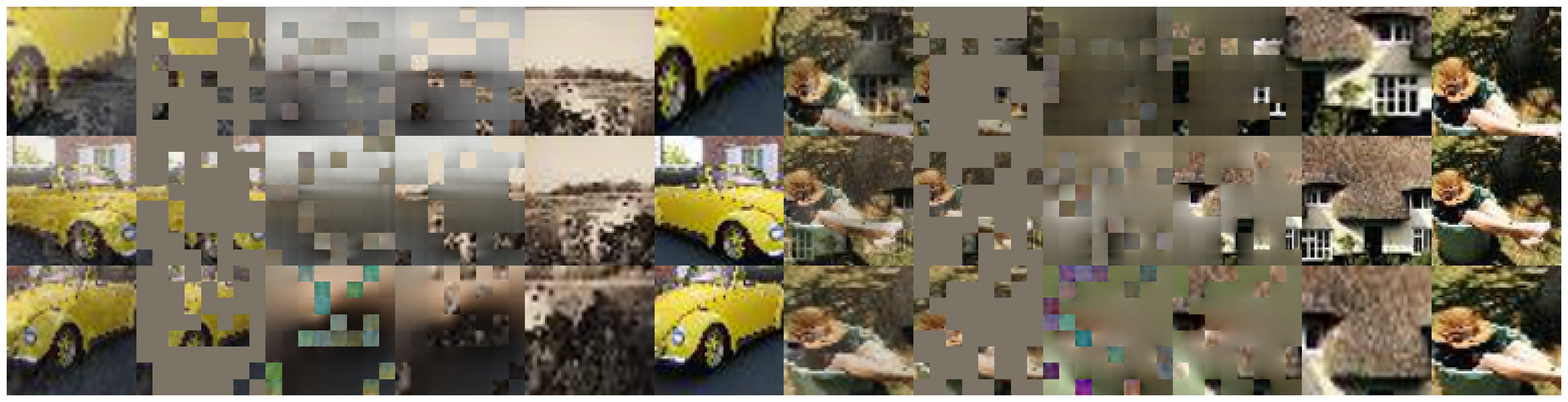}
    \vspace{-0.15in}
    \caption{Different ViT backbones (tiny, small, and base) on Tiny-ImageNet. Reconstruction quality is moderately improved when a larger backbone is used.}
    \label{fig:arch}
    
\end{figure}

\noindent\textbf{ViT architecture.} In our non-ImageNet datasets, we adopt smaller ViT backbones that generally follow~\cite{deit}. The central implementation of linear separation happens between the MAE encoder and decoder, with a linear projection layer for each branch of reconstruction. A shared decoder is used to reconstruct both images. A qualitative evaluation of different ViT sizes on Tiny-ImageNet is displayed in Fig.~\ref{fig:arch}. The perceptive difference is not large, and generally, ViT-small/tiny are sufficient for non-ImageNet datasets. 

\noindent\textbf{Pre-training.} The default setting for pre-training is listed in Tab.~\ref{fig:pretrain}. On ImageNet-1K, we strictly use MAE's specifications. For better classification performance, we use normalized pixels~\cite{he2022masked} and a masking ratio of 0.75. For better visual reconstructions, we use a lower masking ratio of 0.5 without normalizing target pixels. In CIFAR-10/100, and Tiny-ImageNet, reconstruct ordinary pixels.

\noindent\textbf{Semantics-enhanced sampling.} The default settings for our semantics-enhanced mixtures are listed in Tab.~\ref{fig:semconf}. We modified the dataloader to mix, within a mini-batch, $\bm r$ percent of samples that have homogenous classes and $1-\bm r$ percent that have different ones.

\noindent\textbf{Classification.} For the classification task, we provide the detailed settings of our finetuning process in Tab.~\ref{fig:ftconf} and linear evaluation process in Tab.~\ref{fig:linevalconf}.

\begin{algorithm}[h]
\SetAlgoLined
    \PyComment{$\alpha$: mixture ratio alpha} \\
    \PyComment{b = hyperparameter for the Beta Distribution} \\
    \PyComment{c = coefficient for balancing distillation loss and reconstruction loss} \\
    \PyComment{$E_i$, $E_m$: i-MAE encoder, vanilla encoder for distillation} \\
    \PyComment{$D_i$: i-MAE decoder} \\
    \PyComment{$\mathcal{F}_1$, $\mathcal{F}_2$: linear decomposition layer 1, 2 } \\
    \PyCode{\textcolor{red}{def} forward(img):} \\
    \Indp  
        \PyCode{a = random.beta(b, b)} \\ 
        \PyCode{x1, x2 = img, perm(img)} \PyComment{Perm can be random permutation for inner batch mix, or semantics-enhanced sampling} \\
        \PyCode{mix = $\alpha$ * x1 + (1-$\alpha$) * x2} \\
        \PyComment{subordinate and dominant image} \\ 
        \PyCode{x1, x2 = x1, x2 \textcolor{blue}{if} $\alpha$ \textcolor{magenta}{<} 0.5 \textcolor{blue}{else} x2, x1} \\ 
        
        \PyCode{w1, w2  = $E_m$(x1), $E_m$(x2) } \\
        \PyCode{z1, z2  = $\mathcal{F}_1$($E_i$(mix)), $\mathcal{F}_2$($E_i$(mix))} \\
        \PyCode{}\\
        \PyCode{loss  = c * distill(w1,w2,z1,z2) + recon(z1, z2, x1, x2)} \\

        \PyCode{\textcolor{red}{return} loss} \\
    \Indm
    \PyCode{}\\
    
    \PyCode{\textcolor{red}{def} distill(w1, w2, z1, z2):} \\
    \Indp  
        \PyCode{loss = l2\_loss(z1, x1)} \\ 
        \PyCode{loss += l2\_loss(z2, x2)} \\ 

        \PyCode{\textcolor{red}{return} loss} \\
    \Indm
    \PyCode{}\\
    \PyComment{$D_i$: i-MAE decoder} \\
    \PyComment{norm\_pix: normalization function on pixel of each masked patch as the target} \\

    \PyCode{\textcolor{red}{def} recon(z1, z2, x1, x2):} \\
    \Indp  
        \PyCode{t1, t2 = norm\_pix(x1), norm\_pix(x2)} \\ 
        \PyCode{p1, p2 = $D_i$(z1), $D_i$(z2)} \\ 
        \PyCode{loss = MSE(p1, t1) + MSE(p2, t2)} \\ 
        \PyCode{\textcolor{red}{return} loss} \\
    \Indm

\caption{PyTorch-style pseudocode for dual reconstruction targets on i-MAE. }
\label{algo:your-algo}
\end{algorithm}

\section{More Visualizations} \label{vis_app}

\begin{figure}[t]
    \centering
    \includegraphics[width=0.96\linewidth,  keepaspectratio]{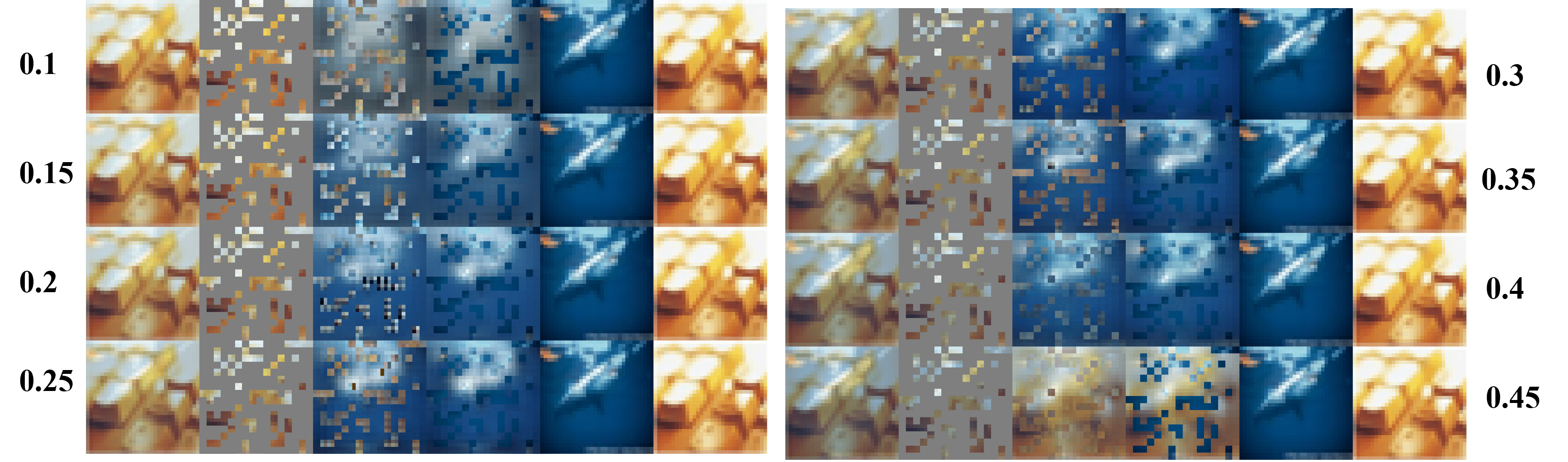}
    \vspace{-0.15in}
    \caption{CIFAR-100 subordinate reconstruction of different ratios marked on the left and right side. Similarly, reconstructions at 0.45 are confused with the dominant image.}
    \label{fig:CIFAR100mix}
    \vspace{0.15in}
    \centering
    \includegraphics[width=0.96\linewidth, height=15cm, keepaspectratio]{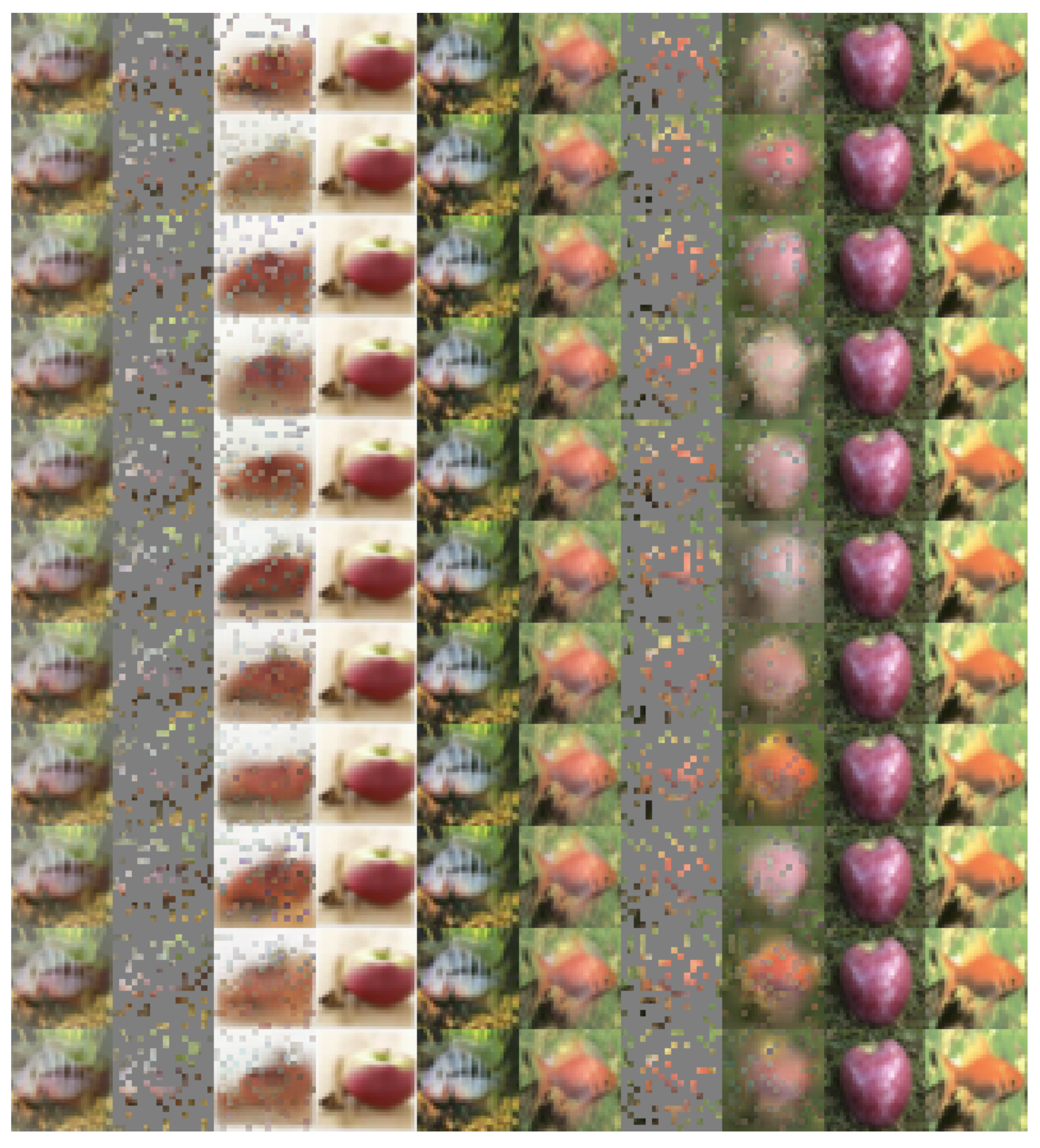}
    \vspace{-0.15in}
    \caption{Uncurated reconstructions of CIFAR-100 validation images using {\em semantics-enhanced mixture} from 0.0 (topmost) to 1.0 (bottom), in 0.1 intervals.}
    \label{fig:my_label}
    
\end{figure}

\begin{figure}[t]
    \centering
    \includegraphics[width=0.98\linewidth, keepaspectratio]{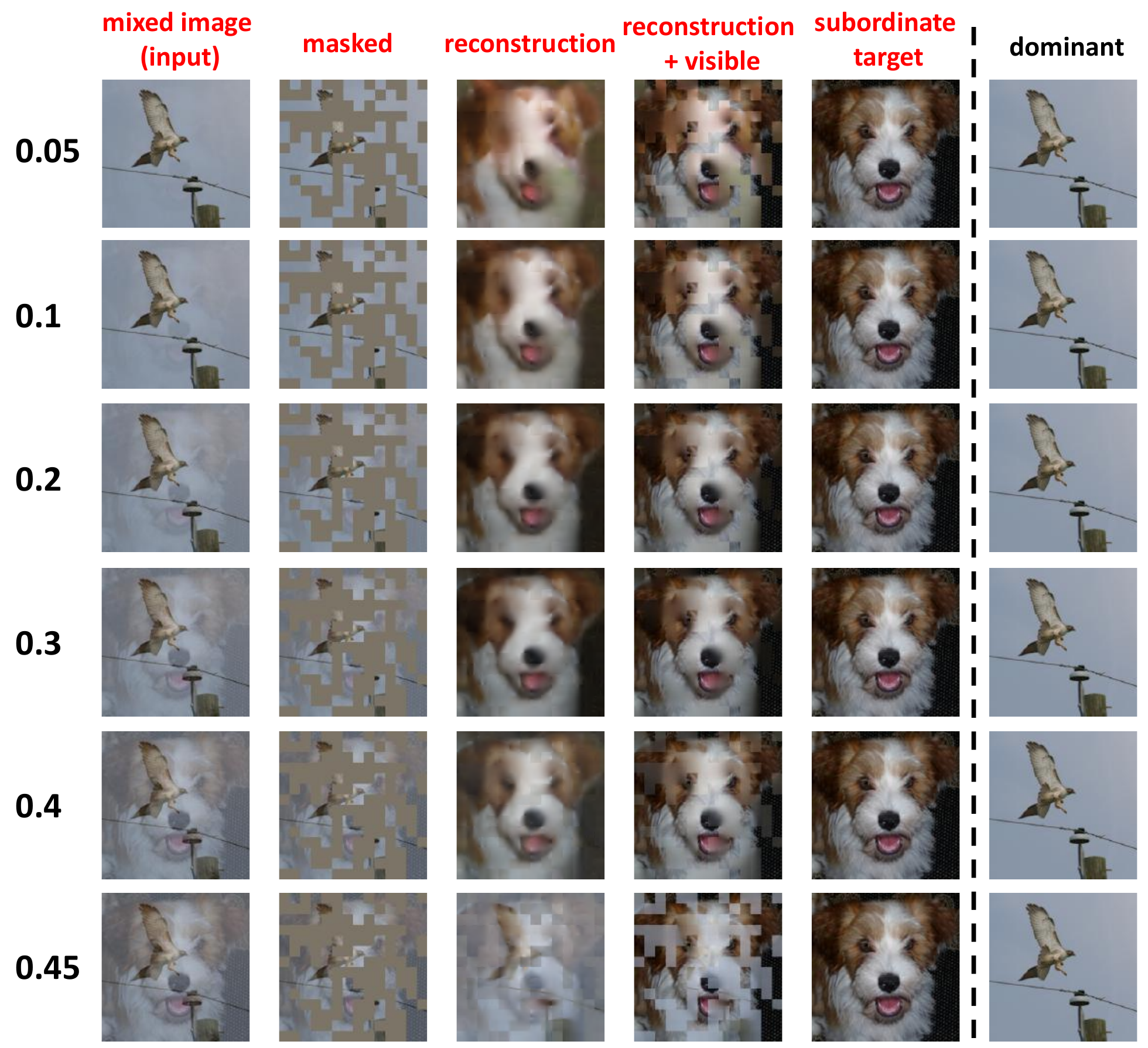}\vspace{0.2in}
    \includegraphics[width=0.98\linewidth, keepaspectratio]{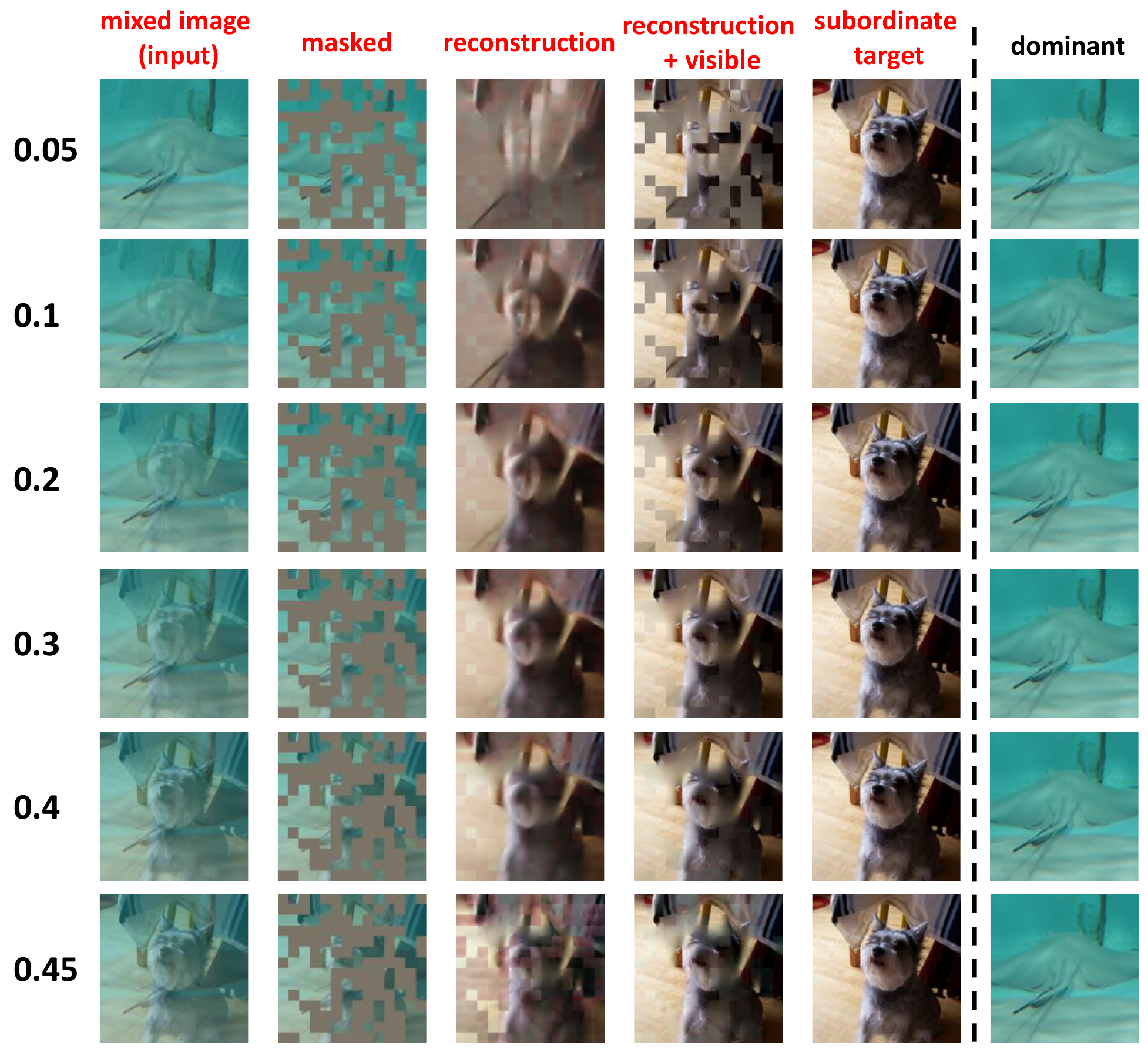}
    \vspace{-0.1in}
    \caption{More reconstructions results of i-MAE on ImageNet-1K {\em validation} images with different mixing coefficients $\bm \alpha$ (listed on the left), 0.5 mask ratio, and with distillation. Visual results are the subordinate reconstructions.}
    \label{fig:ImageNet1}
\end{figure}

\begin{figure}[t]
    \centering
    \includegraphics[width=0.98\linewidth, keepaspectratio]{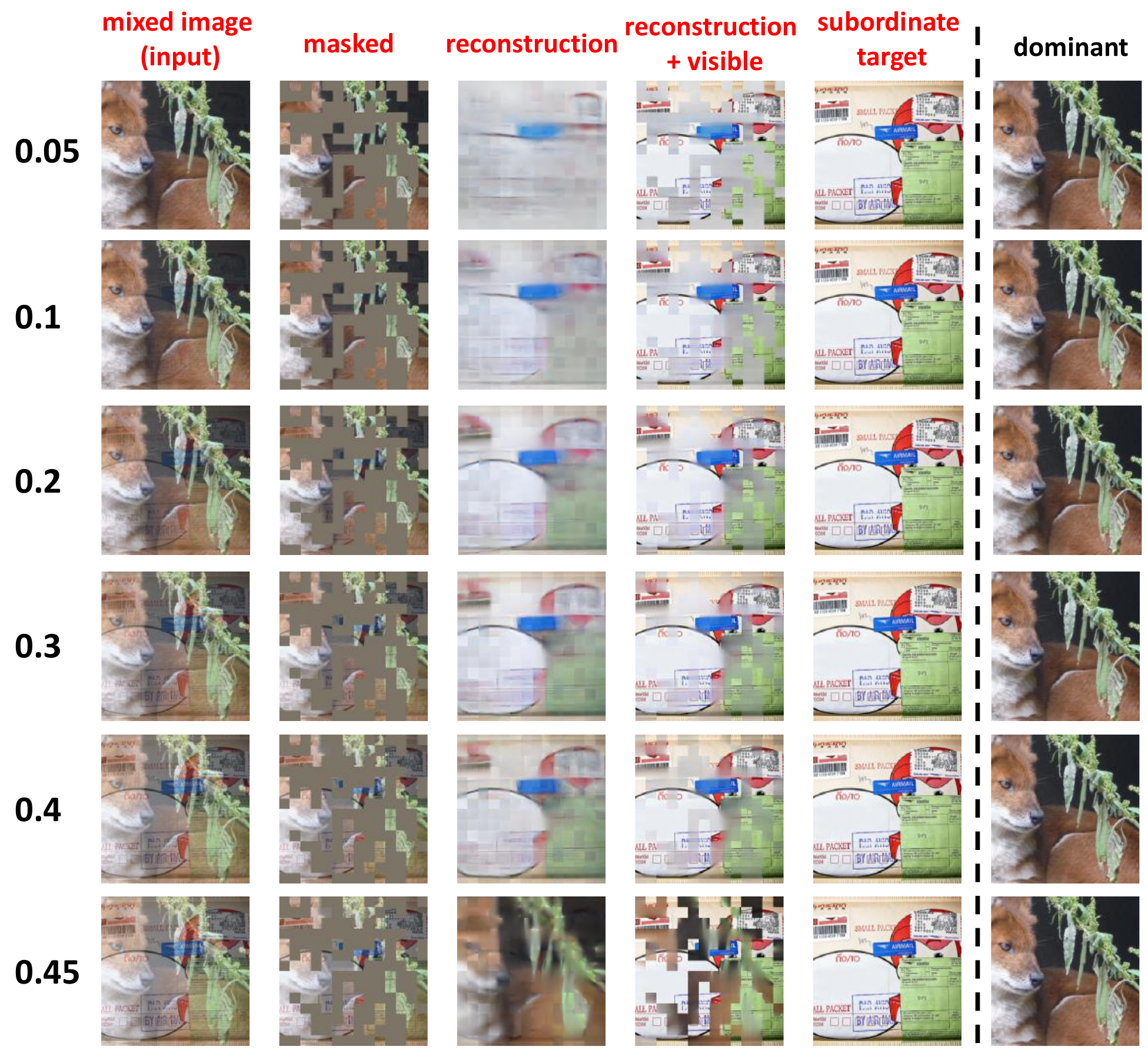}\vspace{0.2in}
    \includegraphics[width=0.98\linewidth, keepaspectratio]{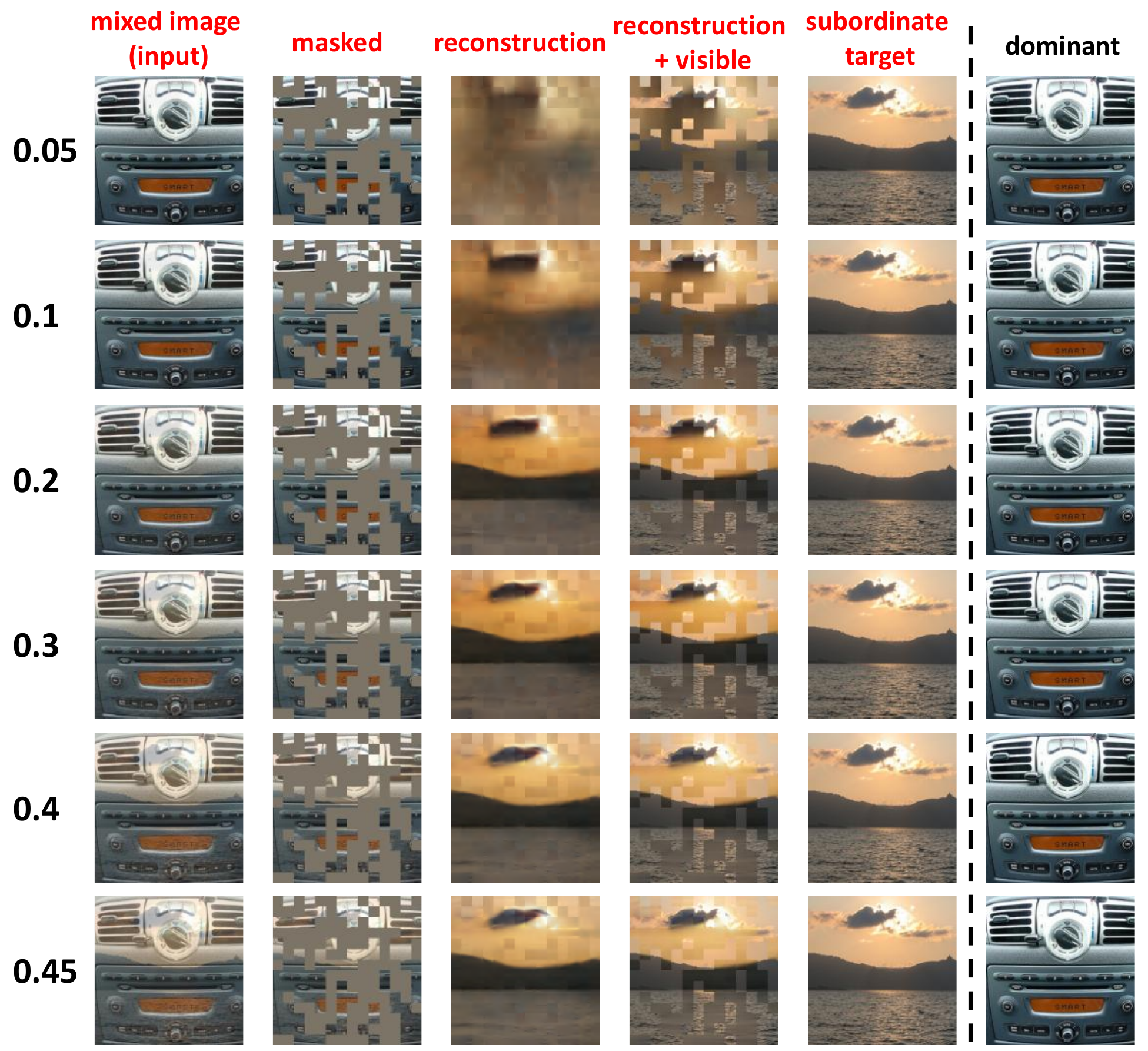}
    \vspace{-0.1in}
    \caption{More reconstructions results of i-MAE on ImageNet-1K {\em validation} images with different mixing coefficients $\bm \alpha$ (listed on the left), 0.5 mask ratio, and with distillation loss. Visual results are the subordinate reconstructions.}
    \label{fig:ImageNet2}
\end{figure}

We provide extra examples of pre-trained i-MAE reconstructing only the subordinate image. Fig.~\ref{fig:CIFAR100mix} are visualizations on CIFAR-100 at mix ratios from 0.1 to 0.45, in 0.05 steps. As depicted in Fig.~\ref{fig:ImageNet1} and Fig.~\ref{fig:ImageNet2}, we produce finer ranges of reconstructions from 0.05 to 0.45. In most cases, mixture rates above 0.4 tend to show features of the dominant image. This observation demonstrates that a low mixture rate can better embed important information separating the subordinate image.  

\section{PyTorch Styled Pseudocode} \label{pseu}
The pseudocode of our mixture and subordinate reconstruction approach is shown in Algorithm~\ref{algo:your-algo}. In our full-fledged i-MAE, we employ two distillation losses for two linear separation branches. ``$\text x2$'' is sampled by a permutation within a mini-batch. Alternatively, we can also employ the semantics-enhanced sampling scheme for it to create $r$ percent of samples from the same class.

\begin{figure}[t]
    \centering
    \includegraphics[width=0.96\linewidth,height=10cm, keepaspectratio]{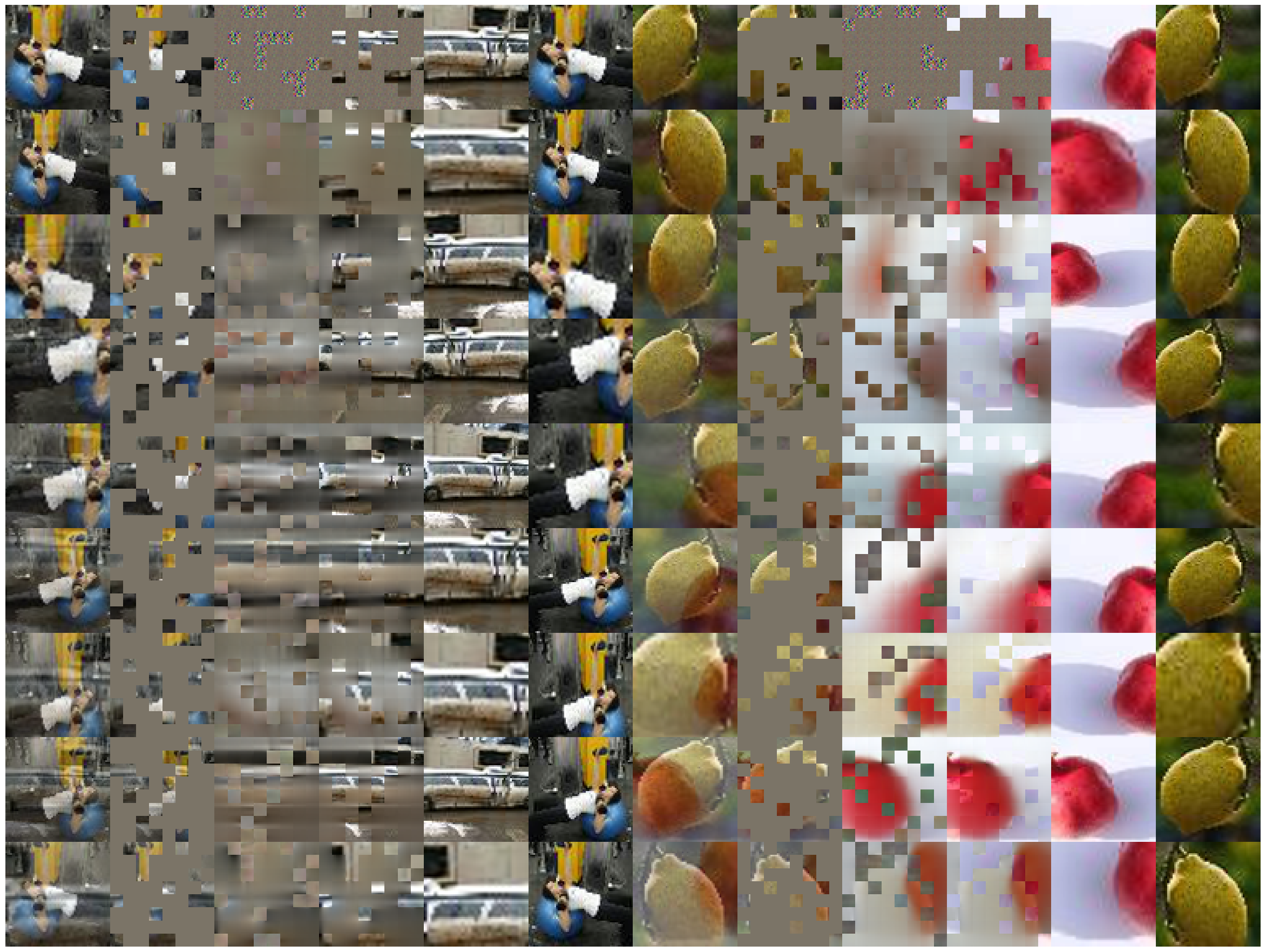}
    \vspace{-0.1in}
    \caption{Uncurated Tiny-ImageNet reconstructions of different mix ratio, from 0.05 to 0.45, subordinate images. }
    \label{fig:tinmix}
    \vspace{0.25in}
    \includegraphics[width=0.96\linewidth, height=15cm, keepaspectratio]{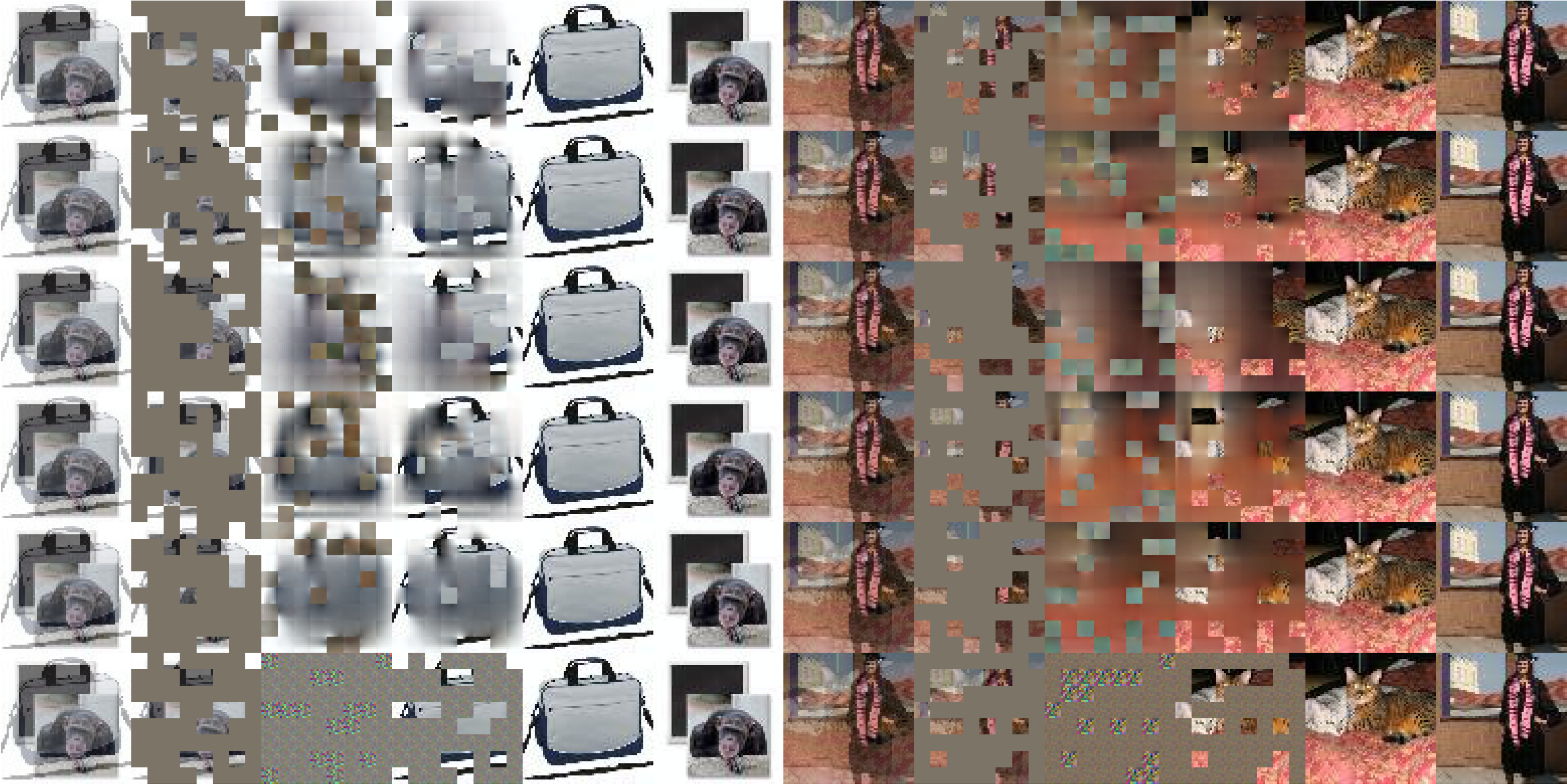}
    \vspace{-0.1in}
    \caption{Visual reconstructions of Tiny-ImageNet validation images using {\em semantics-enhanced mixture} pre-trained i-MAE.}
    \label{fig:sem_mix_tin}
\end{figure}

\begin{table*}[t]
\centering
\small
\begin{tabular}{c | c| c | c  } 
 \hline
 Config & CIFAR-10/100 & Tiny-ImageNet & ImageNet-1K \\ 
 \hline
 base learning rate & 1.5e-4 & 1.5e-4 & 1e-3\\
 \hline
 batch size & 4,096 & 4,096 & 4,096\\ 
 \hline
 Mask Ratio & 0.75 & 0.75 & 0.5  \\
 \hline
  optimizer & AdamW & AdamW & AdamW \\ 
 \hline
 optimizer momentum & 0.9, 0.95 & 0.9, 0.95 & 0.9, 0.95 \\ 
 \hline
 augmentation & None & RandomResizedCrop & RandomResizedCrop \\ 
 \hline
\end{tabular}
\vspace{-0.1in}
\caption{Self-supervised pre-training configurations on CIFAR-10/100, Tiny-ImageNet, ImageNet-1K. For better visualizations, we use a 0.5 mask ratio on ImageNet.}
\label{fig:pretrain}

\vspace{0.1in}
\centering
\begin{tabular}{ c | c | c } 
 \hline
 Config & CIFAR-10/100 & Tiny-ImageNet \\
 \hline
 Object mix range & 0.0, 0.25, 0.5, 1.0 & 0.0, 0.25, 0.5, 1.0 \\ 
 \hline
 Image mix ratio & $\text{Beta}(1.0, 1.0)$ & $\text{Beta}(1.0, 1.0)$  \\ 

 \hline
 base learning rate & 1.5e-4 & 3.5e-4 \\
 \hline
 batch size & 4,096 & 4,096 \\ 
 \hline
 Mask Ratio & 0.75 & 0.75   \\
 \hline
 optimizer & AdamW & AdamW  \\ 
 \hline
 optimizer momentum & 0.9, 0.95 & 0.9, 0.95  \\ 
 \hline
 augmentation & None & RandomResizedCrop  \\ 
 \hline
\end{tabular}
\vspace{-0.1in}
\caption{ Pre-training Configurations of with the {\em semantics-enhanced mixture} scheme.}
\label{fig:semconf}

\vspace{0.1in}

\centering
\begin{tabular}{ c | c | c | c  } 
 \hline
 Config & CIFAR-10/100 & Tiny-ImageNet & ImageNet-1K\\ 
 \hline
 Object mix range & 0.0 - 1.0 & 0.0 - 1.0 & 0.0, 0.25, 0.5, 1.0 \\ 
 \hline
 Image mix ratio & $\text{Beta}(1.0, 1.0)$ & $\text{Beta}(1.0, 1.0)$ &0.8 \\ 

 \hline
 base learning rate & 1e-3 & 1e-3 & 1e-3\\
 \hline
 batch size & 128 & 256 & 1,024\\ 
 \hline
 epochs & 100 & 100 & 25 \\
 \hline
 optimizer & AdamW & AdamW & AdamW \\ 
 \hline
 optimizer momentum & 0.9, 0.999 & 0.9, 0.999 & 0.9, 0.999 \\ 
 \hline
 augmentation & Mixup & Mixup, RandomResizedCrop & Mixup, RandomResizedCrop \\ 
 \hline
\end{tabular}
\vspace{-0.1in}
\caption{Finetune Classification Configurations.}
\label{fig:ftconf}

\vspace{0.1in}

\centering
\begin{tabular}{ c | c | c | c } 
 \hline
 Config & CIFAR-10/100 & Tiny-ImageNet & ImageNet-1K \\ 
 \hline
 Object mix range & 0.0 - 1.0 & 0.0 - 1.0 & 0.0, 0.25, 0.5, 1.0 \\ 
 \hline
 Image mix ratio & 0.35 & 0.35 &0.35 \\ 

 \hline
 base learning rate & 1e-2 & 1e-2 & 1e-2\\
 \hline
 batch size & 128 & 256 & 1,024\\ 
 \hline
 epochs & 200 & 200 & 25 \\
 \hline
 optimizer & SGD & SGD & SGD \\ 
 \hline
 optimizer momentum & 0.9, 0.999 & 0.9, 0.999 & 0.9, 0.999 \\ 
 \hline
 augmentation & None & RandomResizedCrop & RandomResizedCrop \\ 
 \hline
\end{tabular}
\vspace{-0.1in}
\caption{Linear Classification Configurations.}
\label{fig:linevalconf}
\end{table*}

\end{document}